\documentclass[manuscript,screen]{acmart}
\usepackage{graphicx}
\usepackage{amsmath}
\usepackage{amsfonts}
\usepackage{algorithm}
\usepackage{algorithmic}
\usepackage{diagbox}
\usepackage{multirow}
\usepackage{booktabs}
\usepackage{caption}
\usepackage{subfigure}
\usepackage{geometry}
\usepackage{color}
\AtBeginDocument{%
  \providecommand\BibTeX{{%
    \normalfont B\kern-0.5em{\scshape i\kern-0.25em b}\kern-0.8em\TeX}}}

\begin{document}

\title{Enhancing Heterogeneous Knowledge Graph Completion with a Novel GAT-based Approach}


\author{Wanxu Wei}
\email{handsome\_wwx@sjtu.edu.cn}
\affiliation{%
  \institution{Shanghai Jiao Tong University}
  \streetaddress{Dongchuan Road 800}
  \city{Shanghai}
  \country{China}
  \postcode{200240}
}

\author{Yitong Song}
\email{yitong\_song@sjtu.edu.cn}
\affiliation{%
  \institution{Shanghai Jiao Tong University}
  \streetaddress{Dongchuan Road 800}
  \city{Shanghai}
  \country{China}
  \postcode{200240}
}

\author{Bin Yao}
\email{yaobin@cs.sjtu.edu.cn}
\authornote{Corresponding author: Bin Yao, email: yaobin@cs.sjtu.edu.cn}
\affiliation{%
  \institution{Shanghai Jiao Tong University}
  \streetaddress{Dongchuan Road 800}
  \city{Shanghai}
  \country{China}
  \postcode{200240}
}


\begin{abstract}
  Knowledge graphs (KGs) play a vital role in enhancing search results and recommendation systems. With the rapid increase in the size of the KGs, they are becoming inaccuracy and incomplete. This problem can be solved by the knowledge graph completion methods, of which graph attention network (GAT)-based methods stand out since their superior performance. However, existing GAT-based knowledge graph completion methods often suffer from overfitting issues when dealing with heterogeneous knowledge graphs, primarily due to the unbalanced number of samples. Additionally, these methods demonstrate poor performance in predicting the tail (head) entity that shares the same relation and head (tail) entity with others. To solve these problems, we propose GATH, a novel \underline{\textbf{GAT}}-based method designed for \underline{\textbf{H}}eterogeneous KGs. GATH incorporates two separate attention network modules that work synergistically to predict the missing entities. We also introduce novel encoding and feature transformation approaches, enabling the robust performance of GATH in scenarios with imbalanced samples. Comprehensive experiments are conducted to evaluate the GATH's performance. Compared with the existing SOTA GAT-based model on Hits@10 and MRR metrics, our model improves performance by 5.2\% and 5.2\% on the FB15K-237 dataset, and by 4.5\% and 14.6\% on the WN18RR dataset, respectively.
\end{abstract}




\maketitle

\section{Introduction}
\label{intro}
Knowledge graphs (KGs) store entities and their relations as triplets. Each triplet, which is also called fact, can be denoted as $(h, r, t)$, where $h$ and $t$ are the head and tail entities, and $r$ is their relation (e.g., the head entity $h$ is "Kobe", the relation $r$ is "Career", and the tail entity $t$ is "Basketball Player"). Brimming with vast amounts of facts, KGs are widely employed to enhance the search results~\cite{yasunaga2021qa, huang2019knowledge, do2022developing} and refine the recommendation results~\cite{wang2019kgat, wang2021learning, hui2022personalized}. However, as KGs expand rapidly in size, the facts it contains may become inaccurate and incomplete, hindering the ability to provide satisfactory search and recommendation results. 

The knowledge graph completion (KGC) technique is dedicated to completing incomplete triples in KGs. Currently, several studies on knowledge graph completion focus on link prediction, which predicts missing values in a possible triplet. Link prediction is usually defined as, given a query triplet $(?, r, t)$ or $(h, r, ?)$, where $?$ represents the missing value that needs to be predicted. For instance, given the query $(?, \text{isMarriedTo}, \text{William})$, the goal of link prediction is to predict that the missing value is "Kate". In this paper, we also study the link prediction problem for knowledge graph completion.

Recently, knowledge graph embedding models based on graph neural network (GNN) have emerged as a successful method for the link prediction task. 
These approaches offer a comprehensive and nuanced representation of the knowledge graph by capturing the semantics of individual triplets as well as the structural information of the whole KG.
Among all GNN-based methods, GAT-based approaches (e.g., GAT~\cite{velivckovic2017graph}, r-GAT~\cite{busbridge2019relational}, KAME~\cite{jiang2021kernel}, DisenKGAT~\cite{wu2021disenkgat}) exhibit superior performance. This is due to the fact that GAT-based approaches assign different weights or importance to each neighbor during the information aggregation process. This allows the model to prioritize learning from important neighbors while diminishing the influence of unimportant connections. 
While significant progress has been made with GAT-based methods, further improvements are still needed for heterogeneous KGs. Heterogeneous KGs consist of diverse types of entities and relations, which provide a better representation of complex and varied real-world data. However, it becomes more challenging to predict entities and relations in such graphs using GAT-based methods.

\begin{figure}[t]
	\centering
	\subfigure[Case 1: The heterogeneous knowledge graph with unbalanced number of relations.] {\includegraphics[width=0.42\textwidth]{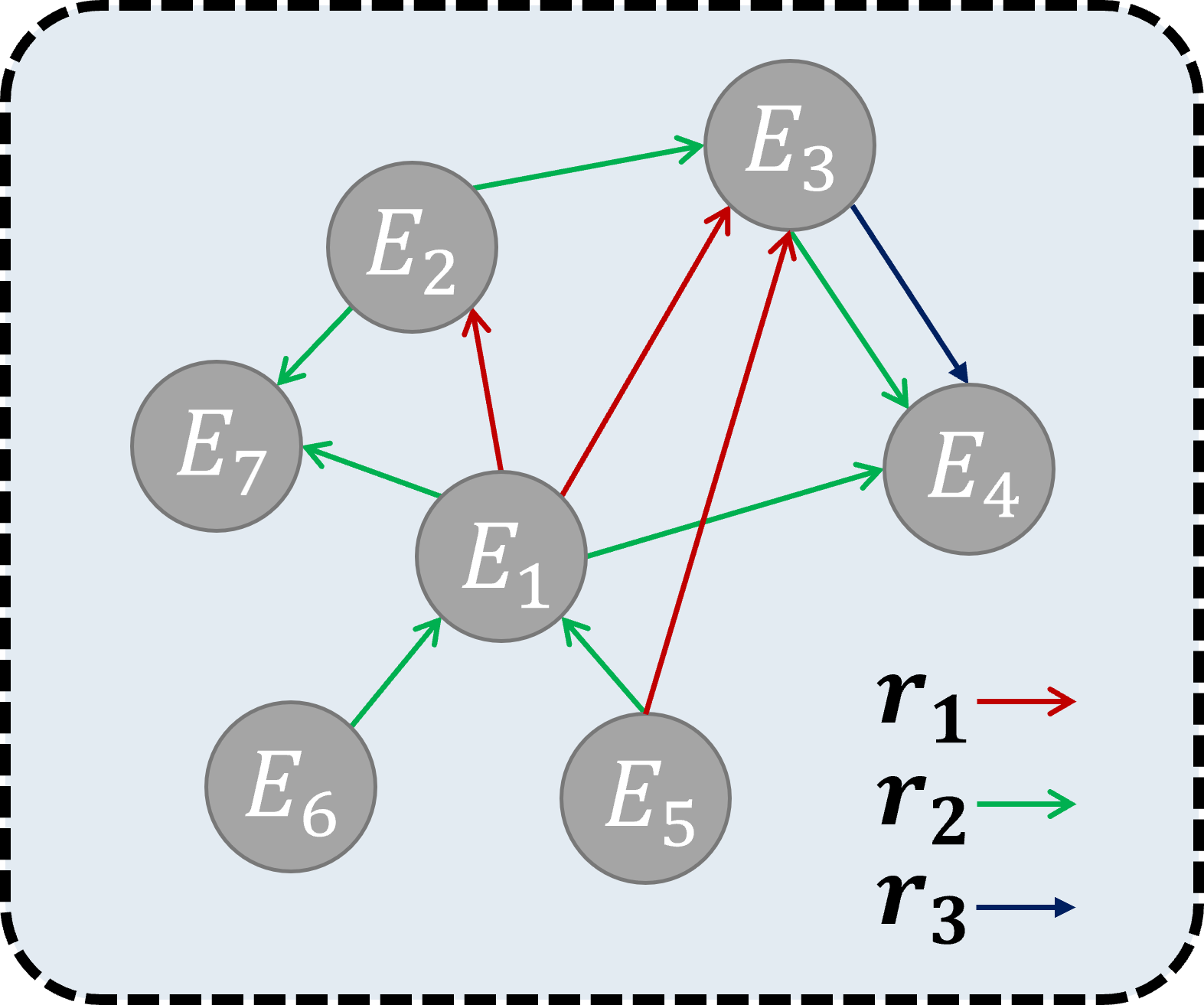}}
	\subfigure[Case 2: The heterogeneous knowledge graph with the entities sharing the same relation with others.] {\includegraphics[width=0.42\textwidth]{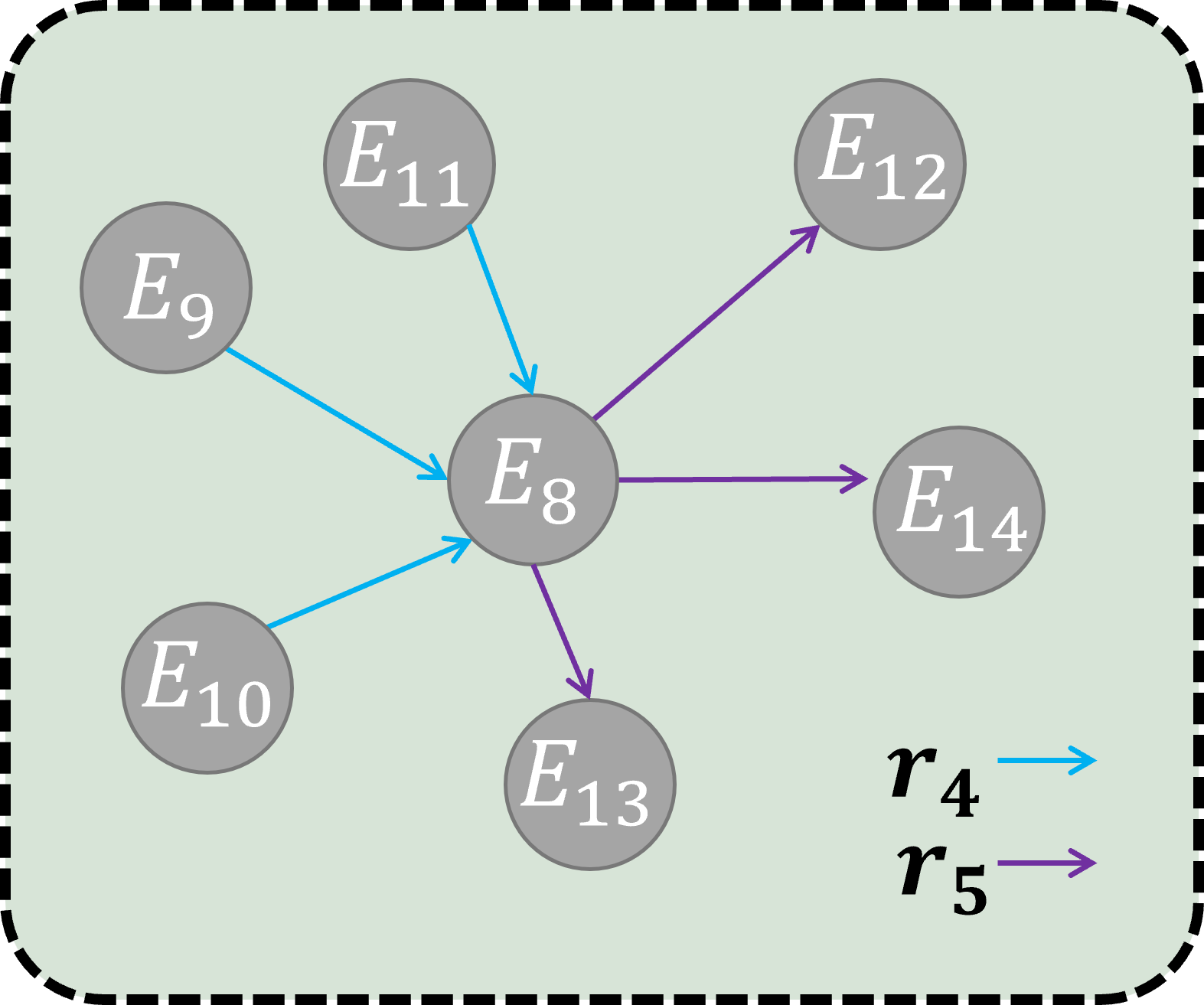}}
	\caption{The examples of heterogeneous knowledge graphs, where the circle represents the entity, labeled as $E_x$ and $x$ is the number. The arrows indicate the relations and different color arrows indicate different types of relations. The relation is noted as $r_x$. }
	\label{heterogeneous graph}
\end{figure}

\textbf{Challenge 1: Existing GAT-based approaches overfit in sparse entity and relation predictions.} 
\textcolor{black}{In deep learning, overfitting is a fundamental problem characterized by a model that can match the training set with high accuracy but performs poorly when modeling other data. \cite{zhang2021understanding} mentions that when the number of model parameters exceeds the amount of information in the training set. the model tends to memorize the training data rather than learn the underlying patterns, resulting in overfitting.}
The distribution of information in the real world tends to be skewed, resulting in some entities and relations appearing infrequently. For example, in Fig.~\ref{heterogeneous graph}(a), only one triple contains node $E_6$ and relation $r_3$. We call these entities and relations with low frequency "sparse entities" and "sparse relations", respectively. 
\textcolor{black}{Existing GAT-based models often have a large number of parameters, which can lead to overfitting when modeling sparse entities and relations due to the parameter quantity exceeding the number of data points. } 
Therefore, existing GAT-based methods suffer from performance degradation due to the presence of sparse entities and relations. For example, in Fig.~\ref{heterogeneous graph}(a), only one triple contains $r_3$. Given the head entity $E_3$ and relation $r_3$, the model is likely to predict the entity $E_4$, whereas the other potential tail entities are masked or overlooked. This is because overfitting occurs when using a large number of parameters to model sparse relations $r_3$ with less information.
As another example, there only exists one triple containing $E_6$, i.e., $(E_6, r_2, E_1)$. And $E_5$ is also related to $E_1$ by $r_2$. Assuming $r_2$ as a $Teammate\_of$ relation, there exists a fact $(E_6, r_2, E_5)$ in the real world but not in the KG. In this case, existing models fail to predict the entity $E_5$ when provided with the head entity $E_6$ and relation $r_2$. This failure is primarily due to insufficient embedded information for $E_6$.

\begin{figure*}[t]
  \centering
  \includegraphics[width=\linewidth]{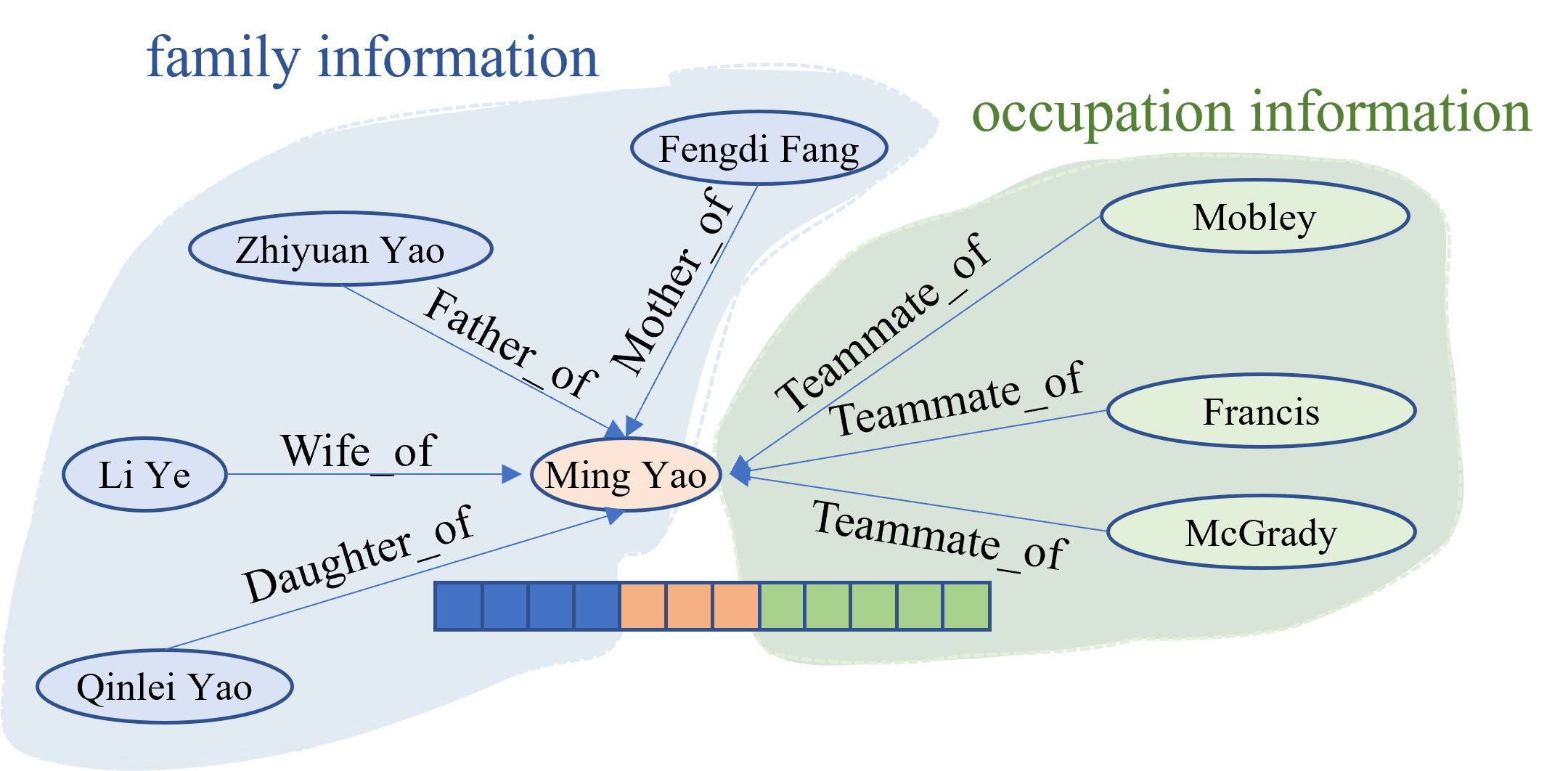}
  \caption{Entity information is represented by a combination of information in different scenarios. In the feature vector, different dimensions correspond to different scenes. A relation is a concrete representation of a scene.}
  \label{figure: feature in different scenarios}
\end{figure*}

\textbf{Challenge 2: Existing GAT-based approaches demonstrate poor performance in predicting the tail (head) entity that shares the same relation and head (tail) entity with others.} 
In heterogeneous KGs, it is quite common to encounter scenarios where different tail (head) entities share the same relation with a particular head (tail) entity, as depicted in Fig.~\ref{heterogeneous graph}(b).
In such cases, existing GAT-based approaches assign weights to entities based on relations.
\textcolor{black}{This will directly lead to the following two limitation: }
\textcolor{black}{1) Some information about neighboring entities may be lost. Relations may prioritize specific information about entities while disregarding other information. For example, consider the example illustrated in Fig.~\ref{figure: feature in different scenarios}, where the entity of a basketball player, $Ming\ Yao$, encompasses various aspects of information such as family information and occupation information. However, the relation $Teammate\_of$ primarily focuses on the occupation information while neglecting other aspects.}
\textcolor{black}{2) Another limitation lies in the inability to capture the varying emphasis that the head entity places on different tail entities within the same relation. Even through the same relation, the head entity will have different inclinations towards neighboring entities. For example, as shown in Fig.~\ref{figure: feature in different scenarios}, $Ming\ Yao$ may have varying inclinations towards different teammates because of information unrelated to occupation information, such as personality and family. This difference cannot be explained if we only consider the relation $Teammate\_of$.}

In this paper, we propose a novel end-to-end GAT-based method for heterogeneous knowledge graph completion, aiming to overcome the limitations of existing GAT-based methods. 
\textcolor{black}{In response to challenge 1, that is, the model is overfitting on sparse entities and relations due to a large number of parameters, our proposed solution is to reduce model parameters.}
\textcolor{black}{Specifically, we first use \textbf{reducing features} to reduce the number of parameters representing relations. Instead of using matrices, we represent relations using embedding vectors. 
Additionally, reducing parameters may limit the model's ability to model relations with rich information. To address this, we introduce \textbf{weight sharing} to enhance the feature extraction capabilities of relations. We use the same attention projection matrices for all relations, which have a global perspective of KG.
By applying these two methods, we reduce the parameters for representing relations from $2nDF$ to $nD+2DF$, where $n$ represents the number of relations, $D$ represents the entity embedding dimension, and $F$ represents the dimension of query or key vectors. In summary, our approach significantly reduces model parameters and effectively mitigates the problem of overfitting.}

\textcolor{black}{Another challenge of existing GAT-based models, as stated in Challenge 2, is that they cannot effectively predict tail (head) entities that share head (tail) entities and relations with other entities.}
This challenge arises from the models' failure to distinguish the importance of entities independently of the relations.
\textcolor{black}{To address this challenge, in addition to calculating attention scores between entities based on relations, we also introduce a novel entity-specific attention network.}
\textcolor{black}{The key to addressing challenge 2 is to address its two limitations, i.e. 1) the loss of partial information in neighboring entities and 2) the inability to represent the varying importance of the head entity to different tail entities connected to the same relation.}
\textcolor{black}{For the first limitation, the additionally introduced entity-specific attention network is capable of directly calculating attention scores for the head entities and the tail entities. Specifically, we linearly transform the head entities into query vectors and the tail entities into key vectors. Both are then fed into a single-layer feedforward neural network to obtain inter-entity attention scores.}
\textcolor{black}{2) Regarding the second limitation, the newly introduced attention network can produce more differentiated attention scores. Specifically, we perform a Hadamard product operation on the query vector and key vector to obtain an intermediate vector. The intermediate vector is used to calculate the attention score. As a result, tail entities that are similar to the head entity get higher attention scores, while dissimilar tail entities receive lower attention scores.}

In conclusion, our contributions are summarized as follows.
\begin{itemize}
\item We introduce GATH, a novel \underline{\textbf{GAT}}-based method specifically designed for \underline{\textbf{H}}eterogeneous knowledge graph completion. GATH comprises two key components: an entity-specific attention network module and an entity-relation joint attention network module. These components work together to predict the missing entities.
\item To address the issue of model overfitting on heterogeneous KGs, we propose a novel encoding and feature transformation method. This method enables the robust performance of GATH in scenarios with sparse samples by effectively reducing the number of model parameters.
\item We conduct a comprehensive evaluation of GATH on the FB15K-237 and WN18RR datasets. Results show that GATH has superior performance than other competitors on various evaluation metrics, including mean reciprocal ranking (MRR), mean ranking (MR), and Hits@n. Compared with the existing SOTA GAT-based model on Hits@10 and MRR metrics, our model improves performance by 5.2\% and 5.2\% on the FB15K-237 dataset, and by 4.5\% and 14.6\% on the WN18RR dataset, respectively.
\end{itemize}

\section{Related work}
Knowledge graph completion methods can be divided into translation-based models, semantic matching-based models, GCN-based models, GAT-based models, and language model-based models, as described below.

\paragraph{Translation-based models} TransE~\cite{bordes2013translating} is a pioneering work of translation-based model. Inspired by word2vec~\cite{mikolov2013distributed}, TransE utilizes the translation invariance of word vectors. It maps the head entity, relation, and tail entity respectively to the dense vectors $h$, $r$, $t$ in a low-dimensional space, and then adjusts $h$, $r$, $t$ so that $h+r\approx t$. Subsequent models, such as TransH~\cite{wang2014knowledge}, TransR~\cite{lin2015learning}, TransD~\cite{ji2015knowledge}, etc., proposed novel methods of relational translation, increasing the complexity of models and improving their performance. These models are also known as \textit{translation models} or \textit{additive models}. However, translation-based models have limited consideration of semantic information and struggle to handle one-to-many or many-to-many connections effectively. Despite their ease of expansion, these models have inherent limitations in capturing complex relations.

\paragraph{Semantic matching-based models} As opposed to translation-based models, semantic matching-based models evaluate the rationality of facts by matching the potential semantics in embedded entities and relations. These models utilize scoring functions to quantify the semantic similarity between entities. RESCAL~\cite{nickel2011three} declares a matrix for each relation and uses the bilinear function to calculate the rationality of triples. DistMult~\cite{yang2014embedding} simplifies RESCAL by replacing the relation-specific matrix with a diagonal matrix. ComplEx~\cite{trouillon2016complex} extends DistMult to the complex space and is the first model to introduce complex number embeddings for knowledge graphs. However, although semantic matching-based models overcome the limitations of translation models by effectively capturing the symmetry or asymmetry of relations, they often struggle to efficiently model complex relational patterns. These models typically require a large number of parameters, leading to potential memory inefficiencies.

\paragraph{GCN-based models} GCN-based models enhance the vanilla GCN for knowledge graph embedding through the incorporation of 1) relation-specific linear transformations, 2) weighted aggregation, and 3) relation representation transformations. Among these, RGCN~\cite{schlichtkrull2018modeling} is the first work that introduces GCN to knowledge graph embedding. RGCN aggregates neighbor information through a relation-specific matrix. After that, SACN~\cite{shang2019end} assigns static weights to each relation type. CompGCN~\cite{vashishth2019composition} jointly embeds entities and relations and uses combination operators to model their interaction. However, GCN-based models often exhibit limited flexibility and require loading the entire knowledge graph into memory, leading to scalability issues and posing challenges when applied to large knowledge graphs.

\paragraph{GAT-based models} GAT-based models achieve state-of-the-art performance in knowledge graph embedding by dynamically modeling the interactions between entities and relations. In contrast to GCN, GAT~\cite{velivckovic2017graph} introduces attention mechanisms to dynamically adjust node weights. Based on GAT, RGAT~\cite{busbridge2019relational} utilizes self-attention to derive relation-related weights for relation features. It combines the inter-entity attention results from GAT to obtain the final outcome. DisenKGAT~\cite{wu2021disenkgat} assumes that entities consist of multiple independent factors and leverages attention mechanisms within each sub-representation space to aggregate neighbor information. HRAN~\cite{li2021learning} incorporates an entity-level encoder to generate neighborhood features for each relation type and a relation-level encoder to capture their relative importance. MRGAT~\cite{dai2022mrgat} utilizes two matrices for each relation type, embedding the head and tail entities into the query and key vectors for attention calculation. However, these models often face challenges such as overfitting on sparse data and limited scalability when dealing with a large number of rapidly increasing relations.

\paragraph{Language model-based models} Knowledge graph text has been successfully leveraged to pre-train embeddings in language model-based models. KG-BERT~\cite{yao2019kg} is the first model to apply BERT~\cite{devlin2018bert}, a pre-training language model widely used in various Natural Language Processing (NLP) tasks, to knowledge graph embedding. KG-BERT offers two variants for knowledge graph completion: KG-BERT (variant a) models triples using their textual descriptions, while KG-BERT (variant b) predicts relations using only the textual descriptions of the head and tail entities. On the other hand, LMKE~\cite{wang2022language} utilizes a language model to generate embeddings for both entities and relations in the knowledge graph. Both KG-BERT and LMKE can effectively use large amounts of text data to pre-train embeddings for knowledge graph entities and relations. However, it is essential to note that the reliance on text data limits the application scenarios of these models.

\section{Background}
\subsection{Knowledge graph}

\paragraph{Knowledge Graph} A knowledge graph (KG) can be denoted as $\mathcal{G=(E,R,T)}$, where $\mathcal{E}$ represents the set of entities, $\mathcal{R}$ represents the set of relations, $\mathcal{T \subseteq  E\times R\times E}$ represents the set of triplets.

\paragraph{Knowledge Graph Embedding} A KG $\mathcal{G}$ can be embedded into a low-dimensional vector space. In other words, a knowledge graph embedding is defined as,
$\mathcal{KGE}=\{(\mathbf{h} ,\mathbf{r} ,\mathbf{t} )|\mathbf{h},\mathbf{t}\in \mathbf{H},\mathbf{r}\in\mathbf{R} \}$, where $\mathbf{H}$ and $\mathbf{R}$ represent the embedding sets of entities and relations which are $n_e\times d_e$ and $n_r \times d_r$ dimensions respectively. $n_e$ and $n_r$ represent the number of entities and relations respectively. $d_e$ and $d_r$ represent the embedding dimensions of entities and relations respectively.

\subsection{Attention mechanism}
\paragraph{Self-Attention Mechanism} Transformer~\cite{vaswani2017attention} introduces the self-attention mechanism, which has been widely adopted in various NLP tasks. Several language models, such as GPT~\cite{radford2018improving} and BERT~\cite{devlin2018bert}, are built upon the self-attention mechanism and have achieved state-of-the-art (SOTA) performance in semantic analysis, machine translation, and other NLP tasks. The self-attention mechanism is responsible for encoding an input sequence consisting of $d_h$-dimensional vectors $\mathbf{H}=[h_1,..., h_N]^T\in \mathbb{R}^{N\times d_h}$, and it can be formulated as follows, 

\begin{equation}
    \text{self\_att}(\mathbf{Q,K,V}) = \text{softmax}(\frac{\mathbf{Q}\mathbf{K}^T}{\sqrt{d_k}})\mathbf{V} \label{selfatt}
\end{equation}
\begin{equation}
    \mathbf{Q=HW_q}, \mathbf{K=HW_k}, \mathbf{V=HW_v}
\end{equation}
where $\mathbf{W_q, W_k\in \mathbb{R}}^{d_h\times d_k}$, $\mathbf{W_v\in \mathbb{R}}^{d_h\times d_v}$ project the input sequence $\mathbf{H}$ to the query matrix $\mathbf{Q}$, key matrix $\mathbf{K}$ and value matrix $\mathbf{V}$ respectively. $\mathbf{Q}$ and $\mathbf{K}$ are multiplied and divided by $\sqrt{d_k}$, and the attention score can be obtained by performing softmax normalization. Finally, we can get the output of the self-attention function by multiplying the attention score and $\mathbf{V}$.

\paragraph{Multi-Head Attention}
The self-attention mechanism is limited to capturing features within a single projection space of $\mathbf{H}$, which makes it sensitive to initialization. To overcome this limitation and capture more diverse features, a multi-head attention mechanism is introduced, drawing inspiration from Convolutional Neural Networks (CNNs). The idea is to leverage multiple projection spaces to capture different aspects of the input. Specifically, if we aim to capture features from $M$ projection spaces, the multi-head attention mechanism can be formulated as follows,

\begin{equation}
    \text{Multihead}(\mathbf{H}) = [\text{head}_1; ...; \text{head}_m]\mathbf{W_o}
\end{equation}
\begin{equation}
    \text{head}_m = \text{self\_att}(\mathbf{Q}_m, \mathbf{K}_m, \mathbf{V}_m)
\end{equation}
\begin{equation}
    \mathbf{Q}_m=\mathbf{H}\mathbf{W}_q^m,\mathbf{K}_m=\mathbf{H}\mathbf{W}_k^m,\mathbf{V}_m=\mathbf{H}\mathbf{W}_v^m
\end{equation}
where $\mathbf{W_o}\in \mathbb{R}^{M * d_v\times d_{h'}}$ is the projection matrix used to convert the slices of multiple self-attention outputs to the original dimension. $head_m$ corresponds to the output sequence of the m-th self-attention function. $\mathbf{W}_q^m, \mathbf{W}_k^m\in \mathbb{R}^{d_h\times d_k}$ and $\mathbf{W}_v^m\in \mathbb{R}^{d_h\times d_v}$ respectively project $\mathbf{H}$ to the query matrix $\mathbf{Q}_m$, key matrix $\mathbf{K}_m$ and value matrix $\mathbf{V}_m$ of the m-th attention head, while $m\in {1,...,M}$

\section{Method}
In this section, we will provide a detailed introduction to GATH. We first introduce the overall framework of GATH and then describe each module and its interconnections in detail. Finally, the loss function used in GATH is proposed.

\subsection{Overall Framework}
\begin{figure*}[t]
  \centering
  \includegraphics[width=\linewidth]{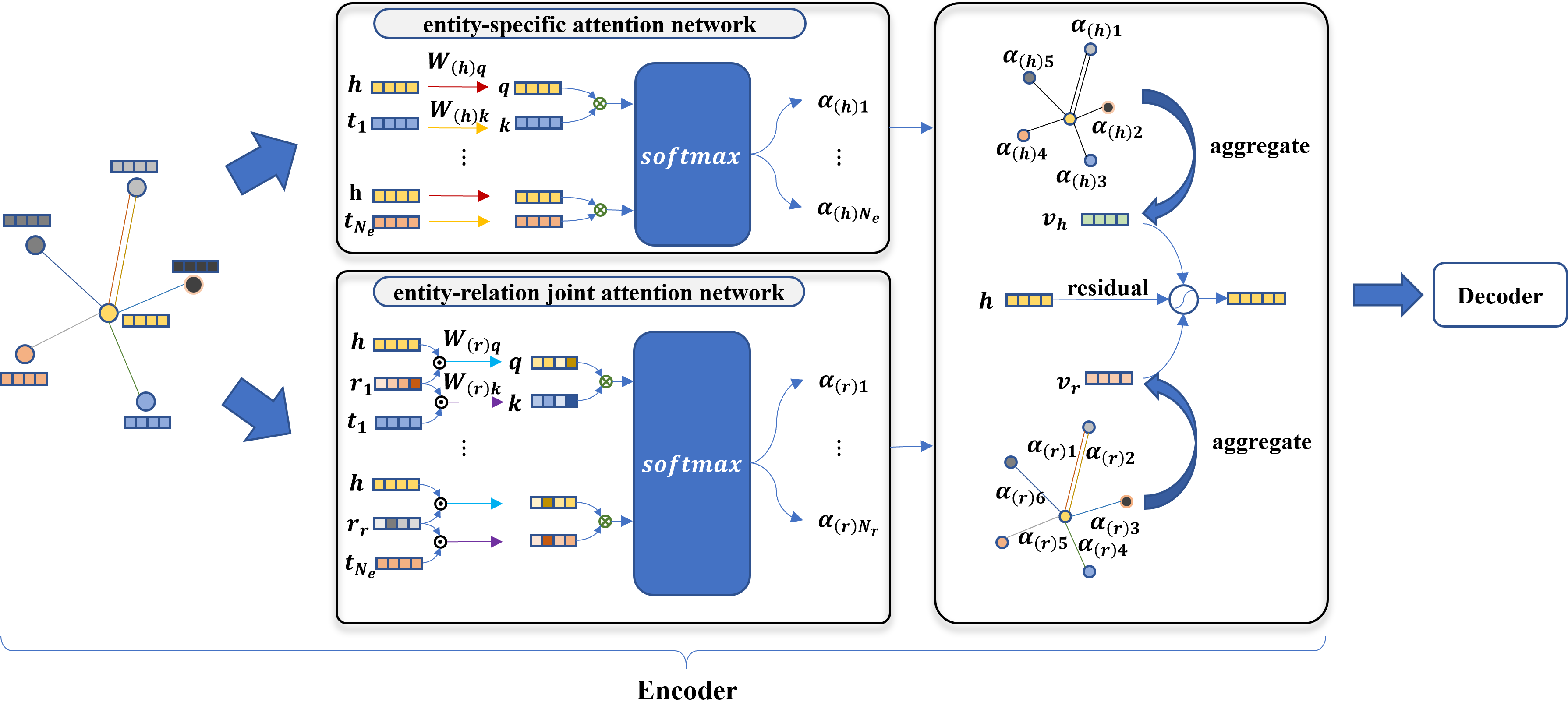}
  \caption{Framework of GATH. GATH consists of two parts: encoder and decoder. 
  In the entity-specific attention network module, an entity's attention score to its neighbors depends on the neighbors themselves. In the entity-relation joint attention network module, an entity's attention score to its neighbors depends on the relation type. Similarly, the aggregation module aggregates information from the entity-specific attention network module and the entity-relation joint attention network module.}
  \label{framework}
\end{figure*}

The architecture of GATH follows the Encoder-Decoder framework, as depicted in Fig.~\ref{framework}. The encoder is responsible for generating embeddings of entities by incorporating both the structural and neighborhood information from the KG. Specifically, the encoder consists of an attention module that calculates attention scores between the central entity and its neighbors, as well as an aggregation module that combines information from all neighboring entities into a single representation. To begin, the attention module computes attention scores via entity pair and relation type, respectively. Subsequently, the aggregation module utilizes these attention scores to aggregate relational and neighborhood information toward the central entity. Finally, the encoder feeds the generated entity embeddings into a decoder. The decoder can be any generic decoder (such as TransE or ConvE) for knowledge graph embedding models. In our case, we employ a decoder based on ConvE with certain enhancements to better capture diverse relational patterns in KGs. Next, we describe each component in detail.

\subsection{Encoder}
The encoder of GATH consists of two primary components: 1) the entity-specific attention network, and 2) the entity-relation joint attention network. These two parts calculate the attention scores of head and tail entities respectively from the entity feature space and the entity-relation joint feature space. They are described separately as follows.

\paragraph{Entity-Specific Attention Network} 
To address the challenge of poor performance in predicting the tail (head) entities that share the same relation and head (tail) entities with others, we introduce an entity-specific attention network that is relation-agnostic. This module aims to capture the intrinsic interaction between a central entity and its neighbor entities, allowing GATH to assign different attention weights to neighbor entities based on their relevance to the central entity regardless of the specific relation involved. By doing so, GATH can efficiently handle the situation shown in Fig.~\ref{heterogeneous graph}(b).

We use $\mathbf{H}^0$ as the initial entity embedding matrix, which is initialized by Gaussian distribution sampling. In $l$-th entity-specific attention network module, the input is $\mathbf{H}^{(l-1)} \in \mathbb{R}^{N_e\times D_h^{(l-1)}}$, where $N_e$ is the number of entities, $l$ represents the number of layers of the encoder, $D_h^{(l-1)}$ represents the dimension of the input vector. We first project the head and tail embeddings into the query and key vectors respectively as follows,
\begin{equation}
    q_{(h)i}^{(l)}=h_i^{(l-1)}\textbf{W}_{(h)q}^{(l)}, k_{(h)j}^{(l)}=h_j^{(l-1)}\textbf{W}_{(h)k}^{(l)}
\end{equation}
where $h_i^{(l-1)}$ and $h_j^{(l-1)}$ denote the embeddings of the i-th node $v_i$ and the j-th node $v_j$ of KG respectively and $v_j$ is in $\mathcal{N}_i$, which is the set of neighbors of $v_i$. $\mathbf{W}_{(h)q}^{(l)}$ and $\mathbf{W}_{(h)k}^{(l)} \in \mathbb{R}^{D_h^{(l-1)}\times d_k}$ are relation-independent learnable projection matrices. $d_k$ is the dimension of the query vector and key vector.

After obtaining the query and key vector, we use a shared attention mechanism when measuring the relation-independent entity importance as follows,
\begin{equation}
    \text{att}^{(l)}_{(h)ij}=f(q_{(h)i}^{(l)} \odot k_{(h)j}^{(l)})
\end{equation}
where $\odot$ means Hadamard product, $f(\cdot)$ represents a feed-forward network capable of capturing entity features with a parameter matrix of $W\in \mathcal{R}^{d_k\times 1}$, followed by a nonlinear function.

Finally, softmax is used to make the importance smoother, i.e.,
\begin{equation}
    \begin{aligned}
    \alpha^{(l)}_{(h)ij}    &=\text{softmax}_j(att^{(l)}_{(h)ij})\\
                            &=\frac{\exp(att^{(l)}_{(h)ij})}{\sum_{v_k\in\mathcal{N}_i}\exp(\text{att}^{(l)}_{(h)ik})}
    \end{aligned}
\end{equation}
where $\alpha^{(l)}_{(h)ij}$ represents the attention score of the central node $v_i$ to its neighbor $v_j$ when no additional relation information is considered. 

\paragraph{Entity-Relation Joint Attention Network} 
In heterogeneous KGs, entities and relations are embedded in different feature spaces, which poses a challenge to effectively capture the joint features of entities and relations. A general approach is to generate a query projection matrix $W_Q\in\mathbb{R}^{D\times F}$ and a key projection matrix $W_K\in\mathbb{R}^{D\times F}$ for each relation, project head, and tail entities as query and key vectors, respectively, and then obtain the attention score, where $D$ represents the dimension of entity embedding and $F$ represents the dimension of query vector and key vector. For a KG with $n$ types of relations, the model needs $2*n*D*F$ space to store relation features. Therefore, this calculation method requires too many parameters for handling numerous relations in large-scale KGs while risking overfitting on sparse relations. Moreover, it is difficult to extend this approach to handle KGs with many different relations due to its high space complexity.
\textcolor{black}{To simplify the parameters representing the relation, we use a $D$-dimensional embedding vector to represent the relation. Furthermore, we make two shared attention projection matrices to map head (tail) entity embeddings transformed by relations to query (key) vectors.}

We assume that each entity is represented by a combination of information in different scenarios. For example, as shown in Fig.~\ref{figure: feature in different scenarios}, a person entity may be represented by a combination of information in various scenarios, such as occupation information and family information. Specifically, in the entity feature space, information in different dimensions corresponds to entity information in different scenarios. A relation is a concrete representation of a scene, so each relation focuses on information under the specific dimensions of the entity feature. We refer to entity features transformed by specific relations as entity-relation joint features. We will compute the attention score based on the entity-relation joint feature.

\textcolor{black}{In practice, in the $D$-dimensional relation embedding that matches the entity embedding dimension, the value under each dimension indicates the degree of attention of the relation to the entity information under the corresponding dimension. In other words, a $D$-dimensional entity embedding can be represented as consisting of $D$ subspaces of dimension $1$. The relation-based entity feature transformation is that the relation enhances or weakens the characteristics of the entity information in each subspace. The following is a detailed demonstration of the calculation process.}

The entity-relation joint attention network internally preserves relation features $\mathbf{R}^{(l)} \in \mathbb{R}^{N_r\times D_h^{(l-1)}}$ and receives entity embeddings $\mathbf{H}^{(l-1)}$, where $N_r$ represents the number of types of relations and $D_h^{(l-1)}$ denotes the dimensionality of each entity embedding. We then obtain the joint features of entities and relations through the following operations,
\begin{equation}
\label{Formula: 9}
    \text{joint}_i^{(l)}=h_i^{(l-1)}\odot r_{r}^{(l)}, 
    \text{joint}_j^{(l)}=h_j^{(l-1)}\odot r_{r}^{(l)}
\end{equation}
where $r_{r}^{(l)}$ represents the embedding of the $r$-th relation, and $(v_i, r_r, v_j)$ is in $\mathcal{N}_i$. $\text{joint}_i^{(l)}$ represents the joint feature of the head entity and relation, and similarly, $\text{joint}_j^{(l)}$ corresponds to the joint feature of the tail entity and relation.

The parameter matrices shared by all relations project the joint features to the query and key vectors. This process can make the information in each dimension of the joint features interact. The attention score will be calculated as follows,
\begin{equation}
\label{Formula: 10}
    q_{(r)i}^{(l)}=\text{joint}_i^{(l)}\textbf{W}_{(r)q}^{(l)}, k_{(r)j}^{(l)}=\text{joint}_j^{(l)}\mathbf{W}_{(r)k}^{(l)}
\end{equation}
\begin{equation}
    \text{att}^{(l)}_{(r)ij}=f(q_{(r)i}^{(l)}, k_{(r)r}^{(l)})
\end{equation}
where $\mathbf{W}_{(r)q}^{(l)}, \mathbf{W}_{(r)k}^{(l)} \in \mathbb{R}^{D_h^{(l-1)}\times d_k}$ are learnable projection matrices shared by all relations. $f(\cdot)$ is a fully connected layer with an output dimension of 1.

In order to make the attention scores more comparable, a softmax operation will be performed,
\begin{equation}
    \begin{aligned}
    \alpha^{(l)}_{(r)ij}&=\text{softmax}_j(att^{(l)}_{(r)ij})\\
                        &=\frac{\exp(\text{att}^{(l)}_{(r)ij})}{\sum_{(v_i, r_r, v_k)\in\mathcal{N}_i}\exp(\text{att}^{(l)}_{(r)ik})}
    \end{aligned}
\end{equation}
where $\alpha^{(l)}_{(r)ij}$ represents the relation-specific head-to-tail attention score.

\paragraph{Analysis of Model Parameters.} 
\textcolor{black}{The model space complexity of GATH is $O(L(N_eD+N_rD+DF))$, while the model space complexity of MRGAT is $O(L(N_eD+N_rDF))$, where $D$ represents the embedding dimension. $N_e = |\mathcal{E}|$ and $N_r = |\mathcal{R}|$ represent the number of entities and relations respectively. $F$ represents the dimension of the query and key vectors. $L$ represents the number of encoder layers. It can be seen that reducing features and weight sharing strategies we proposed significantly reduce the space complexity.} Specifically, 
1) Reduced features: The reduced features are shown in Formula~\ref{Formula: 9}. We assume that the relation is a specific representation in a certain scene, which will enhance or weaken the information of different dimensions in the entity embedding. The parameter amount of each relation feature is $D$. 
2) Weight sharing: If a learnable query projection matrix and key projection matrix are generated for each relation, the parameter amount is $2nDF$. Inspired by the feature map sharing convolution kernel in CNN, we make all the relations share the projection matrix, as shown in Formula~\ref{Formula: 10}. 
Finally, we reduced the relation features from $2nDF$ to $nD+2DF$.

\paragraph{Aggregation}
The aggregation module linearly combines the embeddings of the neighbors according to the attention scores calculated above, and then aggregates this information to generate a new embedding for the central node. For node $v_i$, the formula for aggregating neighborhood information is as follows,
\begin{align}
    h^{(l)}_{(h)i}&=\sum_{v_j\in \mathcal{N}_i}\alpha^{(l)}_{(h)ij}h_j^{(l-1)}\mathbf{W}^{(l)} \\
    h^{(l)}_{(r)i}&=\sum_{(v_i, r_r, v_j)\in \mathcal{N}_i}\alpha^{(l)}_{(r)ij}h_j^{(l-1)}\mathbf{W}^{(l)}
\end{align}
where $\mathbf{W}^{(l)}\in \mathbb{R}^{D^{(l-1)}_h\times d_v}$ is a learnable transformation matrix that projects entity embedding to value vectors. $d_v$ represents the dimension of the value vector. $h^{(l)}_{(h)i}$ and $h^{(l)}_{(r)i}$ denote the embeddings that aggregate the neighborhood from the entity-specific attention network module and the entity-relation joint attention network module, respectively. 

The multi-head attention mechanism enables simultaneous focus on different feature spaces, allowing for the extraction of richer feature information. This has two main advantages. Firstly, it can accelerate model convergence by capturing diverse aspects of the input data. Secondly, it effectively mitigates the sensitivity of the attention mechanism to initialization, enhancing the stability and robustness of the model. Therefore, GATH adopts a multi-head attention mechanism instead of the single-head attention mechanism in order to learn different semantic features more effectively. Assuming there are $M$ attention heads, we use the following formula to concatenate the output of $M$ heads and project them into the entity feature space,
\begin{equation}
    \label{formula:16}
    \hat{h_i}^{(l)} = \text{CONCAT}(h^{(l), 1}_{(h)i}, h^{(l), 1}_{(r)i}, ..., h^{(l), M}_{(h)i}, h^{(l), M}_{(r)i})\mathbf{W}^{(l)}_{o}
\end{equation}
where $h^{(l), m}_{(h)i}$ and $h^{(l), m}_{(r)i}$ denote the embeddings generated by the m-th entity-specific attention network module and entity-relation joint attention network module respectively. $\mathbf{W}^{(l)}_{o}\in \mathbb{R}^{2Md_v\times D_h^{(l)}}$ transforms the multi-head attention output to the entity feature space. $D_h^{(l)}$ denotes the dimension of the output entity embedding of the aggregation module.

Up to this point, the entities have gathered all the neighborhood information. It is worth noting that the information of the entities themselves is not propagated, which may lead to an over-smoothing of the graph, i.e., the entity embeddings tend to be the same. Inspired by ResNet\cite{he2016deep}, we introduce self-loops as residual connections and a learnable parameter $\beta$ for each entity to make information propagation more flexible. The updated formula for entity embedding propagation is as follows,
\begin{equation}
    h^{(l)}_i = \sigma(\hat{h_i}^{(l)}+\beta h_i^{(l-1)}\mathbf{W}^{(l)}_h)
\end{equation}
where $\sigma(\cdot)$ is a non-linear activation function. $\mathbf{W}^{(l)}_h\in \mathbb{R}^{D_h^{(l-1)}\times D_h^{(l)}}$ is propagation matrix.

\begin{algorithm}[!ht]
    \caption{Forward Propagation of Entity and Relation Embedding Matrix in Encoder}
    \label{propagation}
    \begin{algorithmic}[1]
        \REQUIRE{Knowledge Graph $\mathcal{G=(E,R,T)}$\\
        Initial entity embedding $\mathbf{H}\in \mathbb{R}^{N_e\times D_h^{(0)}}$\\
        Encoder layers $L$\\
        Attention heads $M$\\
        Self-loop coefficient $\beta$
        }
        \ENSURE{Generic entity embedding $\mathbf{H}\in \mathbb{R}^{N_e\times D_h^{(L)}}$}
        \FOR{l from 1 to L}
        \FOR{i from 1 to $N_e$}
        \STATE get all neighbors $N_i$
        \FOR{m from 1 to M}
        \FORALL{$(v_i, r_r, v_j) \in N_i$}
        \STATE $\alpha^{(l)}_{(h)ij} \longleftarrow attention(\mathbf{h_i}, \mathbf{h_j})$
        \STATE $\alpha^{(l)}_{(r)ij} \longleftarrow attention(joint(\mathbf{h_i}, \mathbf{r_r}), joint(\mathbf{h_j}, \mathbf{r_r}))$
        \ENDFOR
        \STATE get m-th head output: $h^{(l)}_{(h)i}$, $h^{(l)}_{(r)i}$
        \ENDFOR
        \STATE $\hat{h_i}^{(l)} \longleftarrow Aggregate(M)$
        \STATE $h^{(l)}_i = \sigma(\hat{h_i}^{(l)}+\beta h_i^{(l-1)}\mathbf{W}^{(l)}_h)$
        \ENDFOR
        \ENDFOR
    \end{algorithmic}
\end{algorithm}

\subsection{Decoder}
GATH can adopt any knowledge graph embedding model's decoder, such as TransE or ConvE, etc. Our decoder is implemented based on ConvE, which uses 2D convolutions to model KGs and has better performance in predicting highly connected entities. In addition, inspired by semantic matching-based models, our decoder additionally introduces relation-based entity feature transformation to accurately evaluate the plausibility of triples.

Our encoder generates entity embeddings $\mathbf{H}$ and feeds them into the decoder. The decoder defines the relation embedding matrix $\mathbf{R}\in \mathbb{R}^{N_r\times D_h^{(l)}}$, initialized with Gaussian distribution sampling. The $STACK$ operation, which will concatenate a sequence of vectors on the first dimension, is performed on the entity embeddings, relation embeddings, and their transformed features to obtain the input. Then, we feed this input into a convolutional neural network to make predictions as shown below,
\begin{equation}
\begin{split}
    \text{input} = \text{STACK}(\mathbf{h}, \mathbf{r}, \mathbf{h\odot r})
\end{split}
\end{equation}
\begin{equation}
    g(h,r,t)=\sigma(vec((\overline{\text{input}}\ast \omega))\mathbf{W})\mathbf{t}
\end{equation}
where $\overline{\cdot}$ converts each $D_h$-dimensional vector into a matrix with dimensions $d_w$ and $d_h$ where $D_h = d_w\times d_h$.

In the feed-forward stage, firstly, each vector in the input is reshaped into a 2D matrix. They are then fed into a 2D convolutional layer with kernel $\omega$, which will produce a feature map tensor $\mathcal{F}\in\mathbb{R}^{c\times m\times n}$. $\mathcal{F}$ contains $c$ feature maps with dimensions $m$ and $n$. Secondly, the $vec(\cdot)$ operation converts the feature map $\mathcal{F}$ into a vector $v_1\in\mathbb{R}^{cmn}$. Then the vector $v_1$ is projected into the entity feature space by $\mathbf{W}\in\mathbb{R}^{cmn\times D_h}$. Thirdly, the nonlinear transformation will be performed to get $v_2\in\mathbb{R}^{D_h}$ followed by computing the inner product of $v_2$ and the tail entity embedding $t$ as the score. Finally, the score is normalized to obtain predicted probabilities as shown,
\begin{equation}
    p(h,r,t)=\text{sigmoid}(g(h,r,t))
\end{equation}

\subsection{Loss function}

\textcolor{black}{If the input triple $(h, r, t)$ is valid in KG, then we want the predicted probability $p(h,r,t)$ to be $1$. Otherwise, we want the predicted probability to be $0$\cite{dai2022mrgat}. So the loss function is defined as follows,}
\begin{equation}
    \begin{aligned}
    \mathcal{L}(h,r,t)
    =&-\frac{1}{N}\sum^{N}_{i=1}(y(h,r,t_i)\cdot \log(p(h,r,t_i))+\\
    &(1-y(h,r,t_i))\cdot \log(1-p(h,r,t_i)))
    \end{aligned}
\end{equation}
in which
\begin{equation}
    y(h,r,t_i)=\left\{\begin{matrix}
 1\quad \text{if}\ (h,r,t_i)\in \mathcal{T}\\
 0\quad \text{if}\ (h,r,t_i)\notin \mathcal{T}
\end{matrix}\right.
\end{equation}
where $y(h,r,t_i)$ indicates whether the triplet $(h, r, t_i)$ is a positive triplet, and its value is \{0,1\}. $N$ represents the total number of tail nodes to be predicted. $\mathcal{T}$ represents the set of valid triples.

\section{Experiment}

\subsection{Datasets}

Most of the knowledge graph embedding models and knowledge graph completion models\cite{shang2019end, interacte2020, feng2021novel, dai2022mrgat, yao2019kg, wu2021disenkgat} are developed on the FB15K-237 and WN18RR datasets. Likewise, these two datasets will also be used to evaluate the performance of GATH. A short introduction to these two datasets is given below.

\paragraph{FB15K-237} FB15K-237\cite{toutanova2015observed} is a dataset for link prediction with 14541 entities, 237 relations, and 310,116 triples. It is the subset of FB15K\cite{bordes2013translating}. In FB15K, many triples are inverses that cause leakage from the training to testing and validation splits. And the inverse relations are removed in FB15K-237.

\paragraph{WN18RR} WN18RR\cite{dettmers2018convolutional} is a subset of WN18\cite{bordes2013translating}. The WN18 also suffers from test leakage, similar to FB15K. Therefore, WN18RR has revised WN18 and deletes the inverse relations. In all, WN18RR contains more than 40,000 entities, 11 relations, and more than 90,000 triples. Compared with FB15K-237, WN18RR has more entities, but the complex connections are reduced a lot.

\begin{table}[!t]
    \centering
    \caption{Dataset Statistics}
    \begin{tabular}{|c|c|c|}\hline
        Dataset & FB15K-237 & WN18RR \\\hline
        Entities & 14541 & 40943 \\
        Relations & 237 & 11 \\
        Triplets & 310,116 & 93,004 \\
        Train Triplets & 272115 & 86835 \\
        Val Triplets & 17535 & 3035 \\
        Test Triplets & 20466 & 3134 \\\hline
    \end{tabular}
    \label{tab:Dataset Statistics}
\end{table}

\subsection{Baselines}
In order to demonstrate the superiority of GATH in knowledge graph completion, we mainly conduct comprehensive comparisons with the following 5 categories of knowledge graph embedding models: 1) translation-based models, 2) semantic matching-based models, 3) GCN-based models, 4) GAT-based models and 5) decoder-only models. The calculation process related to the baselines is shown in Table~\ref{tab:baseline introduction}.

\paragraph{Translation-based models}
Translation-based models use distance functions as scoring functions. We use TransE\cite{bordes2013translating} and RotatE\cite{sun2019rotate} as our baselines. This is because TransE is a pioneering work based on translation, while RotatE views relation as the rotation of entity in complex number space and significantly improves the performance of embedded models.

\paragraph{Semantic matching-based models}
Semantic matching-based models usually use similarity functions as scoring functions, such as DistMult \cite{yang2014embedding} and ComplEx \cite{trouillon2016complex}. Similarly, DistMult and ComplEx measure the similarity of triples in real and complex number spaces, respectively. We choose DistMult because it can succinctly obtain high-quality entity embeddings, while ComplEx introduces complex entity embeddings making the model performance greatly improved.

\paragraph{GCN-based models}
GCN-based models use static relation attributes to aggregate neighbor information. 
Among them, we have chosen SACN \cite{shang2019end} and NoGE \cite{Nguyen2022NoGE} as baselines. SACN aggregates neighbor information by assigning static weights to relations, while NoGE aggregates neighbor information with the help of dual quaternion. Although SACN obtains the optimal result, NoGE introduces quaternion to improve the performance. Therefore, SACN and NoGE were chosen as the baselines.

\paragraph{GAT-based models}
GAT-based models combine graph neural networks and attention to dynamically aggregate neighbor information. 
We chose RGAT \cite{busbridge2019relational}, DisenKGAT \cite{wu2021disenkgat}, and MRGAT \cite{dai2022mrgat} as baselines. RGAT extends the original GAT to take relation into account, while DisenKGAT assumes that entity information is composed of multiple sub-feature information and aggregates neighbor information in different subspaces. MRGAT transforms head and tail entities into attention components via relation-specific projection matrices. The reason we chose these three baselines is that RGAT expands the original GAT and considers the relations in KG, greatly improving the performance of the embedding models; DisenKGAT has a similar idea to ours, where entities are composed of features from multiple subspaces together; and MRGAT is currently the state-of-the-art model.

\paragraph{Decoder-only models}

Decoder-only models work on the initial embedding of entities, such as ConvE \cite{dettmers2018convolutional}, Conv\_TransE \cite{shang2019end}, InteractE \cite{interacte2020}, and CTKGC \cite{feng2021novel}. ConvE is similar to Conv\_TransE. Specifically, ConvE reshapes the input while Conv\_TransE does not perform additional processing on the input to maintain translation invariance. InteractE enhances the expressive power of models by increasing the possible interactions between embeddings. CTKGC relies on the fact that the multiplication of the head entity and relation is approximately equal to the tail entity. We choose these baselines for the following reasons. ConvE and InteractE respectively achieved their best performance at the time and were chosen as the baseline by us. ConvE was pioneering work in applying CNN as a decoder, while Conv\_TransE is the decoder of SACN. As a decoder-only model, CTKGC has achieved excellent results in the current knowledge graph embedding models.

\begin{table*}[!t]
    \centering
    \caption{The feature propagation and scoring function of all baselines. $||\cdot||_p$ represents the p-norm of the vector, and p is usually set to 2. $<\cdot>$ represents the dot product of vectors. $\overline{\cdot}$ denotes conjugate for complex vectors in ComplEx, and 2D reshape for real vectors in ConvE and other models. $\circ$ means Hadamard product. $\ast$ indicates convolution operation. $\sigma$ represents a non-linear activation function, such as tanh. $||_1^K$ means multi-head attention mechanism. $\otimes$ means the Hamilton product. $\bigstar$ denotes depth-wise circular convolution. $\phi_{chk}$ rearranges the input vector sequence and combines the elements at the same position to output. A simple example is:$\phi_{chk}([x_1, x_2, x_3], [y_1, y_2, y_3])=[x_1, y_1, x_2, y_2, x_3, y_3]$. $vec(\cdot)$ means to convert all inputs to a vector. Both RotatE and ComplEx perform embedding in the complex domain.}
    \begin{tabular}{lll}
    \toprule
    \specialrule{0em}{1pt}{1pt} 
        \textbf{Models} & Feature Propagation & Score Function\\
    \midrule 
        TransE      & - & $||h+r-t||_p$\\
        DistMult    & - & $<r, h, t>$\\
        ComplEx     & - & $Re(<r, h, \overline{t}>)$\\
        ConvE       & - & $<\sigma(vec(\sigma([\overline{h}; \overline{r}] \ast \omega))W), t>$\\
        RotatE      & - & $||h\circ r - t||_2$\\
        ConvTransE  & - & $<\sigma(vec(\sigma([h; r] \ast \omega))W), t>$\\
        SACN        & $ h_i^{l+1} = \sigma\left(\sum\limits_{j\in\mathcal{N}_i}\alpha_r^lh_j^lW^l + h_i^lW^l\right) $ & $<\sigma(vec(\sigma([h; r] \ast \omega))W), t>$\\
        RGAT        & $ h_i^{l+1} =\sigma\left([||_1^K\sum\limits_{j\in \mathcal{N}_i}\alpha^{lk}_{ij}h_j^lW_k^l; ||_1^M\sum\limits_{j\in \mathcal{N}_i}g(r)^{lm}h_j^lW_m^l]W^{l+1}\right) $ & - \\
        InteractE   & - & $<\sigma(vec(\sigma(\phi_{chk}(P_k) \bigstar \omega))W), t>$\\
        CTKGC       & - & $<f(vec(\sigma([h^T\otimes r \ast \omega))W), t>$\\
        DisenKGAT   & $h_{i}^{l+1} = \sigma\left(\sum\limits_{(v_j, r_r)\in \mathcal{N}_u}\alpha^k_{(v_i, v_j, r_r)}\phi(h_{v, k}, r, \theta_r)\right) $ & $\sum\limits_k\beta^k<\sigma(vec(\sigma([\overline{h_{k}}; \overline{r}] \ast \omega))W), t_k>$\\ 
        NoGE        & $h_{i, k}^{l+1} = \sigma\left(\sum\limits_{j\in\mathcal{N}_i}a_{i, j}W^k\otimes_d h_j^k\right) $ & $<h, r, t>$ \\
        MRGAT       & $h_i^{l+1}=\sigma\left(h_i^lW^l+||_1^K\sum\limits_{j\in\mathcal{N}_i} a_{ij}^{l.k}h_j^lW_v^{l,k}\right)$ & $<\sigma(vec(\sigma([h; r] \ast \omega))W), t>$\\
        \hline
        \specialrule{0em}{1pt}{1pt}
        GATH        & $h_i^{l+1}=\sigma\left(\beta h_i^lW_h^l+||_1^K\sum\limits_{j\in\mathcal{N}_i}[a_{(h)ij}^{lk}h_jW^l;a_{(r)ij}^{lk}h_jW^l]\right)$ & $<\sigma(vec(\sigma([\overline{h}; \overline{r}; \overline{h_{trans}}] \ast \omega))W), t>$ \\
    \bottomrule 
    \end{tabular}
    \label{tab:baseline introduction}
\end{table*}

\subsection{Experiment setup}
The number of GATH's encoding layer is set to 2, and AdamW\cite{loshchilov2017decoupled} is used as an optimization algorithm for the training of GATH. The above settings are applied to all baselines based on GCN and GAT. At the same time, in order to speed up convergence and prevent model overfitting, GATH also adopts dropout\cite{srivastava2014dropout} and batch normalization\cite{ioffe2015batch} for entity and relation embeddings. Periodic decay of the learning rate is applied to GATH and baselines. A high learning rate makes the loss drop rapidly in the early stage of training and enters a plateau in the middle stage, and finally, a small learning rate makes the model converge slowly. Therefore, we set the initial learning rate to 0.01, and let it decay to 0.985 of the current learning rate every epoch. \emph{batch\_size} and \emph{embedding\_size} are set to 128 and 200 respectively. \textcolor{black}{For baselines that can use the block-diagonal-decomposition regularization scheme\cite{schlichtkrull2018modeling}, we set num\_blocks to $50$.}

The environment of GATH is python3.9+pytorch1.12 and it runs on NVIDIA 4090 Graphics Processing Units. The computation time of GATH for each epoch is about 4.3 minutes and 1.6 minutes for the FB15K-237 and WN18RR datasets respectively.

\subsection{Evaluation indicators}
To evaluate the performance of GATH, we perform the link prediction task on the test set. Specifically, for each triplet $(h, r, t)$ in the test set, we construct the corresponding reverse triplet $(t, r\_reverse, h)$. Each triplet $(e1, r, e2)$ in the constructed test set will be used for prediction. For the decoder, it receives the head entity $e1$ and relation $r$ as input and outputs the probabilities of all tail entities. The prediction result is a list $pred$, whose length is the number of all entities, and the corresponding element $pred[i]$ is the probability of triplet $(e_1, r, e_{v_i})$ in KGs. Since the triplets in the training set and test set will also appear in the prediction list, $pred$ needs to be filtered, that is, the probability of the corresponding triplet is set to 0. Sort $pred$ from high to low to generate $spred$, and the index of $e_2$ in $spred$ is the ranking score for $(e_1, r, e_2)$. The specific metrics include Mean Reciprocal Rank (MRR), Mean Rank (MR) and the percentage of correct entities ranked top 1, 3, and 10 (Hits@1, Hits@3, Hits@10). These metrics are calculated as follows,
\begin{align*}
    \text{MRR}     &= \frac{1}{|S|}\sum_{i=1}^{|S|}\frac{1}{\text{rank}_i} \\
    \text{MR}      &= \frac{1}{|S|}\sum_{i=1}^{|S|}\text{rank}_i \\
    \text{Hits@}n  &= \frac{1}{|S|}\sum_{i=1}^{|S|}I(\text{rank}_i\le n), where\ n = 1, 3, 10
\end{align*}

\section{Result}


\subsection{Overall performance}

The performance of GATH is shown in Table \ref{tab:model performance}. When compared with other existing knowledge graph embedding models, GATH can achieve the best performance in most indicators on FB15K-237 and WN18RR datasets. Next, we will conduct a more in-depth analysis of the results shown in Table \ref{tab:model performance}:

First, we compare decoder-only baselines with our decoder. On FB15K-237, ConvTransE has better performance than other decoder-only baselines. It can be seen that our decoder is 1.7\% higher than ConvTransE on the Hits@10 indicator, 7.3\% on the Hits@1 indicator, and 4.5\% on the MRR indicator. Compared with FB15K-237, WN18RR has a lot less complex connections, but our decoder can still maintain the competition. This demonstrates that relation-based entity feature transformation can improve performance when modeling heterogeneous KGs with complex connections.

Second, our proposed GATH is able to perceive structural information through graph attention networks, improving the quality of entity embeddings. By comparing GATH with our decoder, all indicators have been significantly improved. On FB15K-237, GATH improves Hits@1, Hits@10, and MRR by 15.5\%, 11.2\%, and 13.5\% compared to our decoder. Similarly, on WN18RR, the performance of GATH improves by 7.0\%, 9.4\%, and 7.7\%, respectively. This demonstrates the effectiveness of GATH's encoding layer.

Thirdly, it's noteworthy that our proposed GATH outperforms other GNN-based models, such as SACN and MRGAT, in most evaluation metrics. For example, compared with the best-performing model MRGAT in the GNN-based model, the three indicators (Hits@1, Hits@10, and MRR) of our GATH on FB15K-237 have increased by 5.4\%, 5.2\%, and 5.2\% respectively, and the corresponding increases of 24.2\%, 4.5\%, and 14.6\% on WN18RR. This indicates that GATH is more advanced. Unlike GCN-based models, GATH can adaptively learn the importance between entities to aggregate neighbor information. Furthermore, by exploring the intrinsic attention within entities, GATH provides higher-quality entity embeddings for downstream decoders, compared to most GAT-based models that only consider computing attention based on relational features.

Overall, after a comprehensive comparison with different types of knowledge graph embedding models, we find that our proposed GATH model shows significant competitiveness on all evaluation metrics. On the one hand, as a variant of GNN, GATH can capture the topological information in the graph, thus outperforming traditional knowledge graph embedding models such as TransE, RotatE, and ConvE, which only focus on independent triplets. On the other hand, compared to other GCN-based and GAT-based models, such as SACN, RGAT, NoGE and MRGAT, GATH still exhibits more advanced performance because its encoder can focus on important node information under complex connections within heterogeneous multi-relational graphs.

\begin{table*}[!t]
    \centering
    \caption{Link prediction performance on FB15K-237 and WN18RR datasets}
    \begin{tabular}{lccccccccccc}
    \toprule
    \specialrule{0em}{1pt}{1pt} 
        \textbf{Models} & \multicolumn{5}{c}{\textbf{FB15K-237}} & \textbf{} & \multicolumn{5}{c}{\textbf{WN18RR}} \\\cline{2-6} \cline{8-12}
        \specialrule{0em}{1pt}{1pt} 
        ~ & \multicolumn{3}{c}{Hits@n} & MR & MRR & ~ & \multicolumn{3}{c}{Hits@n} & MR & MRR \\ \cline{2-4} \cline{8-10}
        \specialrule{0em}{1pt}{1pt} 
        ~ & 1 & 3 & 10 & ~ & ~ & ~ & 1 & 3 & 10 & ~ & ~ \\
    \midrule 
        TransE      & 0.105 & 0.24 & 0.425 & 401 & 0.219 & ~ & 0.05 & 0.318 & 0.458 & 5548 & 0.2 \\
        DistMult    & 0.158 & 0.247 & 0.363 & 1105 & 0.226 & ~ & 0.362 & 0.405 & 0.456 & 9249 & 0.392 \\
        ComplEx     & 0.215 & 0.329 & 0.466 & 666 & 0.298 & ~ & 0.386 & 0.437 & 0.493 & 8580 & 0.42 \\
        ConvE       & 0.207 & 0.316 & 0.461 & 362 & 0.29 & ~ & 0.394 & 0.434 & 0.490 & 5554 & 0.425 \\
        RotatE      & 0.205 & 0.317 & 0.457 & 515 & 0.288 & ~ & 0.406 & 0.451 & 0.494 & 7409 & 0.436 \\
        ConvTransE  & 0.204 & 0.32 & 0.466 & 432 & 0.29 & ~ & 0.255 & 0.322 & 0.383 & 7043 & 0.3 \\
        SACN        & 0.239 & 0.361 & 0.5 & 205 & 0.328 & ~ & 0.334 & 0.407 & 0.474 & 6194 & 0.378 \\
        RGAT        & 0.234 & 0.36 & 0.506 & 191 & 0.326 & ~ & 0.083 & 0.278 & 0.432 & 4086 & 0.202 \\
        InteractE   & 0.211 & 0.317 & 0.464 & 575 & 0.296 & ~ & 0.416 & 0.455 & 0.499 & 5281 & 0.444 \\
        CTKGC       & 0.202 & 0.305 & 0.433 & 742 & 0.278 & ~ & 0.423 & 0.464 & 0.506 & 5862 & 0.451 \\
        DisenKGAT   & 0.227 & 0.347 & 0.487 & 242 & 0.314   & ~ & 0.379 & 0.441 & 0.498 & 4823 & 0.421\\ 
        NoGE        & 0.235 & 0.361 & 0.511 & 218 & 0.326 & ~ & 0.067 & 0.36 & 0.47 & 3397 & 0.226 \\
        MRGAT       & 0.24 & 0.356 & 0.501 & \textbf{178} & 0.327 & ~ & 0.343 & 0.437 & 0.514 & \textbf{2615} & 0.404 \\
        \specialrule{0em}{1pt}{1pt}\hline
        \specialrule{0em}{1pt}{1pt}
        our decoder & 0.219 & 0.331 & 0.474 & 296 & 0.303 & ~ & 0.398 & 0.443 & 0.491 & 5964 & 0.43 \\
        GATH        & \textbf{0.253} & \textbf{0.376} & \textbf{0.527} & 192 & \textbf{0.344} & ~ & \textbf{0.426} & \textbf{0.475} & \textbf{0.537} & 3442 & \textbf{0.463} \\
    \bottomrule 
    \end{tabular}
    \label{tab:model performance}
\end{table*}

\subsection{Ablation study}

\begin{table*}[!ht]
    \centering
    \caption{Ablation study. The first line is that the model only uses our decoder for link prediction. The second line is the result of adding an encoder that uses the entity-relation joint attention network. The third line is the complete GATH, with both entity-specific attention network and entity-relation joint attention network added to the encoder.}
    \begin{tabular}{l|ccc|ccc|ccc|cc}
    \toprule
    \specialrule{0em}{1pt}{1pt} 
        \textbf{Models} & \multicolumn{5}{c}{\textbf{FB15K-237}} & \textbf{} & \multicolumn{5}{c}{\textbf{WN18RR}} \\\cline{2-6} \cline{8-12}
        \specialrule{0em}{1pt}{1pt} 
        ~ & \multicolumn{3}{c}{Hits@n} & MR & MRR & ~ & \multicolumn{3}{c}{Hits@n} & MR & MRR \\ \cline{2-4} \cline{8-10}
        \specialrule{0em}{1pt}{1pt} 
        ~ & 1 & 3 & 10 & ~ & ~ & ~ & 1 & 3 & 10 & ~ & ~ \\
    \midrule 
    Decoder-only & 0.219 & 0.331 & 0.474 & 296 & 0.303 & ~ & 0.398 & 0.443 & 0.491 & 5964 & 0.43 \\
    Entity-relation joint attention network-only & 0.253 & 0.377 & 0.524 & 168 & 0.343 & ~ & 0.401 & 0.465 & 0.52 & 3246 & 0.445\\
    GATH    & 0.253 & 0.376 & 0.527 & 192 & 0.344 & ~ & 0.426 & 0.475 & 0.537 & 3442 & 0.463 \\
    \bottomrule 
    \end{tabular}
    \label{tab: attention module}
\end{table*}

Table~\ref{tab: attention module} shows the results of our ablation study. Through observation, we find that: (1) The addition of the entity-specific attention network resulted in an increase in most link prediction indicators, which proves the existence of intrinsic interaction features between entities and the effectiveness of the entity-specific attention network. (2) After considering entity-specific attention network, there was a significant improvement in link prediction results on the WN18RR dataset, where Hits@1, Hits@10, and MRR increased by 6.23\%, 3\%, and 4.04\%, respectively. But on the FB15k-237, the increase is small, and even the Hits@3 indicator has decreased. This is because the types and number of relations in WN18RR are small, so the interaction intensity between entities and relations is smaller than that of FB15k-237, and the interaction between entities is highlighted, which finally makes the improvement of link predictors by entity-specific attention networks on WN18RR more obvious.

\subsection{The performance with different sample sizes}
\paragraph{Performance on different entity sample sizes}

"Graph degree" refers to the number of edges that a node is connected to in a graph. This concept is widely used in graph theory. Similarly, in KGs, the degree is defined as the number of triples containing a node. The degree of a node is related to its sparsity. A node with a small degree is a sparse node with fewer occurrences in KG. Relatively, a node with a higher degree is a dense node.

Based on the degree of the node, we classify it into Sparse Nodes, Moderate Nodes, and Dense Nodes. Among them, Sparse Nodes is a set of nodes with a degree in the range [0, 100], Moderate Nodes is a set of nodes with a degree in the range  (100, 1000], and Dense Nodes is a set of nodes with a degree greater than 1000. We then conduct link prediction tasks on GATH, as well as embedding models: TransE, DistMult, and ConvE, and graph neural network SACN (GCN-based), and MRGAT (GAT-based), which are all classic works or SOTA embedding models in their respective types of models. We report the results in terms of Hits@10 and MRR metrics, as shown in Table \ref{tab:node degree}.

After studying Table~\ref{tab:node degree}, we find that: (1) When the degree of nodes increases, the model performance often improves. (2) On Sparse Nodes, GATH exhibits the best performance. Compared with the state-of-the-art model MRGAT based on GAT, we have introduced feature reduction and entity-specific attention network, making our model perform better on nodes with a lower degree, such as Sparse Nodes and Moderate Nodes. On the one hand, this proves the effectiveness of our work. On the other hand, this indicates that GATH has higher robustness against sparse nodes.

\begin{table*}[!ht]
    \centering
    \caption{Link prediction performance on different entity sample sizes}
    \begin{tabular}{lcccccccccccccc}
    \toprule
    \specialrule{0em}{1pt}{1pt} 
        \textbf{Models} & \multicolumn{2}{c}{\textbf{Sparse Nodes}} &  & \multicolumn{2}{c}{\textbf{Moderate Nodes}} &  & \multicolumn{2}{c}{\textbf{Dense Nodes}}  \\\cline{2-3} \cline{5-6} \cline{8-9} 
        \specialrule{0em}{1pt}{1pt}
        ~ & Hits@10 & MRR & ~ & Hits@10 & MRR & ~ & Hits@10 & MRR & \\
    \midrule 
        TransE      & 0.421 & 0.218 & ~ & 0.458 & 0.216 & ~ & 0.505 & 0.359  \\
        ConvE	    & 0.457	& 0.289	& ~ & 0.516	& 0.306	& ~ & 0.526	& 0.431	 \\
        DistMult    & 0.361 & 0.227 & ~ & 0.382 & 0.212 & ~ & 0.442 & 0.343  \\
        SACN        & 0.457 & 0.194 & ~ & 0.558 & 0.222 & ~ & 0.570 & 0.484  \\
        MRGAT       & 0.490 & 0.318 & ~ & 0.592 & 0.386 & ~ & 0.607 & 0.507  \\
        \hline
        \specialrule{0em}{1pt}{1pt}
        GATH        & 0.511 & 0.334 & ~ & 0.601 & 0.397 & ~ & 0.594 & 0.505  \\
    \bottomrule 
    \end{tabular}
    \label{tab:node degree}
\end{table*}

\paragraph{Performance on different relation sample sizes}

We choose FB15k-237 as the dataset, which has 237 types of relations. We take the number of triples containing a certain relation in the graph as the degree of the relation. Based on the degree of the relation, we classify it into Sparse Relations, Moderate Relations, and Dense Relations. Among them, Sparse Relations is a set of relations with a degree in the range [0, 200], Moderate Relations is a set of relations with a degree in the range  (200, 500], and Dense Relations is a set of relations with a degree greater than 500. The comparison results between GATH and several other baselines are shown in Table~\ref{tab:relation degree}. For each set of relations, we demonstrate Hits@10 And MRR indicators.

Through studying Table~\ref{tab:relation degree}, we find that: (1) several models in Table~\ref{tab:relation degree} include GATH decrease MRR and Hits@10 scores as the degree of the relation increases. (2) Our model achieved optimal performance on all indicators, especially with significant performance improvement in small-degree relations, i.e., Sparse Relations and Moderate Relations. This proves that our strategy of reduced features and weight sharing can effectively model sparse data in KGs.

\begin{table*}[!ht]
    \centering
    \caption{Link prediction performance on different relation sample sizes}
    \begin{tabular}{lcccccccccccccc}
    \toprule
    \specialrule{0em}{1pt}{1pt} 
        \textbf{Models} & \multicolumn{2}{c}{\textbf{Sparse Relations}} &  & \multicolumn{2}{c}{\textbf{Moderate Relations}} &  & \multicolumn{2}{c}{\textbf{Dense Relations}}  \\\cline{2-3} \cline{5-6} \cline{8-9} 
        \specialrule{0em}{1pt}{1pt}
        ~ & Hits@10 & MRR & ~ & Hits@10 & MRR & ~ & Hits@10 & MRR & \\
    \midrule 
        TransE      & 0.585 & 0.379 & ~ & 0.542 & 0.346 & ~ & 0.407 & 0.199  \\
        ConvE	    & 0.639	& 0.450	& ~ & 0.590	& 0.404	& ~ & 0.443	& 0.274	 \\
        DistMult    & 0.505 & 0.379 & ~ & 0.488 & 0.326 & ~ & 0.345 & 0.210  \\
        SACN        & 0.645 & 0.470 & ~ & 0.612 & 0.423 & ~ & 0.485 & 0.311  \\
        MRGAT       & 0.632 & 0.461 & ~ & 0.596 & 0.406 & ~ & 0.487 & 0.312  \\
        \hline
        \specialrule{0em}{1pt}{1pt}
        GATH        & 0.672 & 0.503 & ~ & 0.626 & 0.438 & ~ & 0.503 & 0.325  \\
    \bottomrule 
    \end{tabular}
    \label{tab:relation degree}
\end{table*}

\subsection{Training utilization analysis}
\textcolor{black}{To compare the training speed and space occupation of the models, we have chosen SACN and MRGAT as the baseline models. SACN is the best-performing GCN-based model in the experiment, and similarly, MRGAT is the optimal GAT-based model. Table~\ref{tab: train utils} shows the time required for GATH and the baseline models to train one epoch, as well as the size of each encoder layer during training.}

\textcolor{black}{By studying Table~\ref{tab: train utils}, we find that: 1) Compared to the GAT-based baseline—MRGAT, GATH performs better in both model occupation and training time. This indicates the effectiveness of our improvements to the current GAT-based models. We use embedding vectors instead of matrices to represent relations, significantly reducing model occupation and improving training efficiency. 2) SACN uses static weights to represent relations. Although it uses fewer training resources, its performance in link prediction is much lower compared to MRGAT and GATH.}

\begin{table*}[!ht]
    \centering
    \caption{Link prediction utilization on FB15K-237 and WN18RR}
    \begin{tabular}{lcccccccccccccc}
    \toprule
    \specialrule{0em}{1pt}{1pt} 
        \textbf{Models} & \multicolumn{2}{c}{\textbf{Time(s/epoch)}} &  & \multicolumn{2}{c}{\textbf{Space(GB/Encoder)}}\\\cline{2-3} \cline{5-6}
        \specialrule{0em}{1pt}{1pt}
        ~ & FB15K-237 & WN18RR & ~ & FB15K-237 & WN18RR \\
    \midrule 
        SACN        & 46    & 54    & ~     & 0.5 & 0.2   \\
        MRGAT        & 375   & 131   & ~     & 20 & 8.5 \\
        \specialrule{0em}{1pt}{1pt}\hline
        \specialrule{0em}{1pt}{1pt}
        GATH        & 260 & 97     & ~ & 9.8 & 3.4 \\
    \bottomrule 
    \end{tabular}
    \label{tab: train utils}
\end{table*}

\section{Conclusion and future work}

We first point out that current knowledge graph completion models cannot maintain good performance when modeling sparse data in heterogeneous KG. Afterward, they struggle when the same head (tail) entity and relation are connected to multiple different tail (head) entities in heterogeneous KGs. To address these issues, we propose a graph attention network called GATH. On the one hand, GATH reduces the impact of sparsity on model performance by reducing features and weight sharing. On the other hand, we consider the intrinsic interactions of entities within the entity feature space to select more important neighbors. 
We conduct a comprehensive evaluation of GATH and achieve SOTA performance. We also test Gath with MindSpore and get similar results.

In the next step, we hope to overcome the over-smoothing problem of the graph neural network that has been planted. And we hope to extend the model to the complex number space to obtain further performance improvements.

\begin{acks}
    This work was supported by the National Key Research and Development Program of China 2020YFB1710200, the NSFC 61832017, and the CAAI-Huawei MindSpore Open Fund.
\end{acks}

\bibliographystyle{ACM-Reference-Format}
\bibliography{GATH}
\end{document}